\pdfoutput=1
\documentclass[11pt]{article}

\usepackage{amsmath}

\usepackage[table,xcdraw]{xcolor}
\usepackage{booktabs}

\usepackage{tikz}
\usetikzlibrary{positioning,shapes,arrows,fit,backgrounds,calc,shadows}

\usepackage{listings}

\usepackage{url}

\usepackage{enumitem}

\usepackage{amsthm}
\theoremstyle{plain}

\theoremstyle{definition}

\theoremstyle{remark}

\usepackage{amsfonts, amssymb}

\newcommand\A{\mathcal{A}}
\newcommand\M{\mathcal{M}}
\newcommand\C{\mathcal{C}}

\newcommand\B{\mathcal{B}}

\newcommand\J{\mathcal{J}}

\usepackage{accents}

\newcommand{\dataname}{MEMERAG}
\newcommand\method{\texttt{STEMF}}

\usepackage[colorinlistoftodos,bordercolor=orange,backgroundcolor=orange!20,linecolor=orange, textwidth=0.7in]{todonotes}

\usepackage{algorithm}
\usepackage{algpseudocode}

\usepackage{natbib} 

\usepackage{graphicx}
\usepackage{subfigure}
\usepackage{caption} 
\usepackage{wrapfig}

\usepackage[preprint]{acl}

\usepackage{times}
\usepackage{latexsym}
\usepackage[T1]{fontenc}
\usepackage[utf8]{inputenc}
\usepackage{microtype}
\usepackage{inconsolata}
\usepackage{float}
\usepackage{comment}

\usepackage[capitalize,noabbrev]{cleveref}
\makeatletter
\renewcommand\thanks[1]{\footnotemark\protected@xdef\@thanks{\@thanks
\protect\footnotetext[\the\c@footnote]{#1}}}
\makeatother

	\crefname{hyp}{}{assumption}
	\Crefname{hyp}{}{Assumption}

\def\showcomments{}  
\ifdefined\showcomments
\newcommand{\aam}[1]{\textcolor{green}{$_{Aymen}${[#1]}}}
    \newcommand{\sm}[1]{\textcolor{cyan}{$_{Saab}${[#1]}}}   
    \newcommand{\mf}[1]{\textcolor{blue}{$_{Marcello}${[#1]}}}    
    \newcommand{\ca}[1]{\textcolor{red}{$_{Carlo}${[#1]}}}    
\else
    \newcommand{\aam}[1]{}
    \newcommand{\sm}[1]{}
     \newcommand{\mf}[1]{}
     \newcommand{\ca}[1]{}
\fi

\title{Multilingual Self-Taught Faithfulness Evaluators}

\author{
 \textbf{Carlo Alfano\textsuperscript{1,*}},
 \textbf{Aymen Al Marjani\textsuperscript{2}},
 \textbf{Zeno Jonke\textsuperscript{2}},
\\
 \textbf{Amin Mantrach\textsuperscript{2}},
 \textbf{Saab Mansour\textsuperscript{2}},
 \textbf{Marcello Federico\textsuperscript{2}}
\\ 
 \textsuperscript{1}University of Oxford,
 \textsuperscript{2}Amazon 
\\
 \small{
   \textbf{Correspondence:} {carlo.alfano@stats.ox.ac.uk}
 }
}

\begin{document}
\maketitle 
\begin{abstract}
The growing use of large language models (LLMs) has increased the need for automatic evaluation systems, particularly to address the challenge of information hallucination. Although existing faithfulness evaluation approaches have shown promise, they are predominantly English-focused and often require expensive human-labeled training data for fine-tuning specialized models. As LLMs see increased adoption in multilingual contexts, there is a need for accurate faithfulness evaluators that can operate across languages without extensive labeled data. 
This paper presents \method~(Self-Taught Evaluators for Multilingual Faithfulness), a framework that learns exclusively from synthetic multilingual summarization data while leveraging cross-lingual transfer learning. Through experiments comparing language-specific and mixed-language fine-tuning approaches, we demonstrate a consistent relationship between an LLM's general language capabilities and its performance in language-specific evaluation tasks. Our framework shows improvements over existing baselines, including state-of-the-art English evaluators and machine translation-based approaches.
\end{abstract}

\renewcommand{\thefootnote}{\fnsymbol{footnote}}
\footnotetext[1]{Work done at Amazon.}

\section{Introduction}
The growing application of large language models (LLMs) for text generation tasks across multiple languages has created an urgent need for efficient and reliable evaluation methods. While human evaluation remains the gold standard for assessing LLM outputs, it is expensive, time-consuming, and difficult to scale across different languages. This has led to increasing interest in automated evaluation approaches, particularly using LLMs themselves as evaluators \citep{alpaca_eval, zheng2023judging}. \looseness =-1

Recent work has demonstrated promising results using LLMs to evaluate other LLMs' outputs, especially for tasks like summarization and QA in English \citep{liu-etal-2023-g, song-etal-2024-finesure, wang2024pandalm}. However, extending these evaluation approaches to multiple languages presents several key challenges. First, most existing LLM-as-judge frameworks have been developed and tested primarily on English data. Second, it remains unclear whether LLMs can deliver consistent evaluation quality across different languages, especially for languages with limited training data. Third, the optimal approach for training multilingual LLM evaluators - whether through language-specific fine-tuning, mixed-language training, or other strategies - is not well understood.
In this paper, we address these challenges by developing a framework for training multilingual faithfulness evaluators with synthetic data, using a self-taught approach \cite{wang2024self}. Our key contributions include:

\begin{itemize}[itemsep=0.7pt, topsep=0pt, parsep=0pt, partopsep=0pt, leftmargin=0.4cm]
    \item \textbf{S}elf-\textbf{T}aught \textbf{E}valuators for \textbf{M}ultilingual \textbf{F}aithfulness (\textbf{\method}), a scalable framework for training multilingual faithfulness evaluators using synthetic summarization data. \looseness = -1
    \item Comprehensive experiments across multiple languages and model architectures to identify optimal training strategies.
    \item A multilingual finetuned faithfulness evaluator based on a 9B-parameter LLM that outperforms baselines of similar dimension and matches the performance of larger models.
    \item Comparison to baselines such as out-of-the-box prompting and translation based approaches.
\end{itemize}

Through extensive experimentation, we demonstrate that our framework can effectively improve evaluation performance across languages while identifying important considerations for deploying such systems in practice. Our findings provide valuable insights for researchers and practitioners working on multilingual LLM evaluation.

The rest of this paper is organized as follows. \Cref{sec:rel_works} reviews related works and \Cref{sec:STEMF} presents the \texttt{\method} framework in detail. \Cref{sec:exp} describes our experimental setup, with results and analysis presented in \Cref{sec:res}. 

\section{Related Work}
\label{sec:rel_works}
Our work is related to three main topics present in the literature: LLMs as evaluators, LLMs taught by LLMs, and multilinguality in LLMs.

\vspace{0.1cm}
\paragraph{LLMs as evaluators} Given the cost of acquiring human annotations, the automatic and effective evaluation of LLMs is a crucial challenge in LLM research that has been growing in difficulty with the advancement of LLM capabilities. Traditional metrics for the automatic evaluation of LLMs, such as BLEU and ROUGE, have been shown to have poor correlation with human annotations when evaluating the output of modern LLMs for most tasks \citep{blagec-etal-2022-global}, urging researchers to find innovative solutions. A promising approach to address this challenge consists in using LLMs to evaluate LLMs, either by designing prompts for off-the-shelf instruction tuned models \citep{alpaca_eval, zheng2023judging, saha-etal-2024-branch}, or by fine-tuning LLMs to act as evaluators \citep{kim2024prometheus, wang-etal-2024-interpretable, tang-etal-2024-minicheck}, especially in a self-supervised fashion. \looseness = -1

\paragraph{LLMs taught by LLMs}
The cost of human annotations affects the fine-tuning of LLMs as well, and researcher have devised methods to fine-tune LLMs using LLMs~\cite{lakew2017improving, pan-etal-2024-automatically}. \citet{yuan2024self} showed that it is possible to fine-tune an LLM using rewards given by the LLM itself. Building on the same idea, \cite{wu2024meta} showed that allowing the LLM to evaluate its own judgments and using the feedback to improve its judgments skills leads to larger improvements than the original strategy. As to evaluators, \citet{wang2024self} and \cite{tang-etal-2024-minicheck} have built LLM-based evaluators using contrastive synthetic annotated data. In particular, they have shown that constructing a dataset by asking an LLM to generate good and bad responses to a set of prompts, and then fine-tuning an LLM on the synthetic dataset, leads to state-of-the-art evaluators. Most existing works on this topic focus on the English language, highlighting a gap in the literature regarding the evaluation of multilingual LLMs evaluators, which we plan to fill. \looseness = -1
 
\paragraph{Multilinguality in LLMs}
In terms of multilingual LLMs, designing training strategies that allow generalization or transfer to different languages is crucial to address the lack of high-quality data in many languages. \citet{artetxe-etal-2020-cross} and later \citet{chen2023improving} propose to transfer an LLM to a new language by freezing all the parameters of the LLM except the embedding, which is reinitialized and re-learned through pretraining on the target language. \citet{zhao2024llama} show that, if provided with sufficient data (1B tokens), these strategies are not necessary and standard pretraining and fine-tuning are sufficient. \citet{bacciu-etal-2024-dantellm} show that performing SFT with QLora on Mistral 7B-Instruct-v0.2 on several Italian datasets increases the performance of the LLM on Italian benchmarks. Besides training, recent works \citep{wendler-etal-2024-llamas, tang-etal-2024-language, zhao2024large} have studied how LLMs handle multilingualism and observed that internal layers of the LLM deal with abstract concepts and task solving, while external layers address the input and output languages. \looseness = -1

Finally, we note that \citet{bavaresco2024llms} and \citet{tang-etal-2024-tofueval} provide evidence that LLMs are not yet ready to systematically replace human judgments, as their agreement with human preferences varies largely across tasks and is particularly lacking on dialogue data. 

\begin{figure*}[t]
    \centering
    \includegraphics[width=1\linewidth]{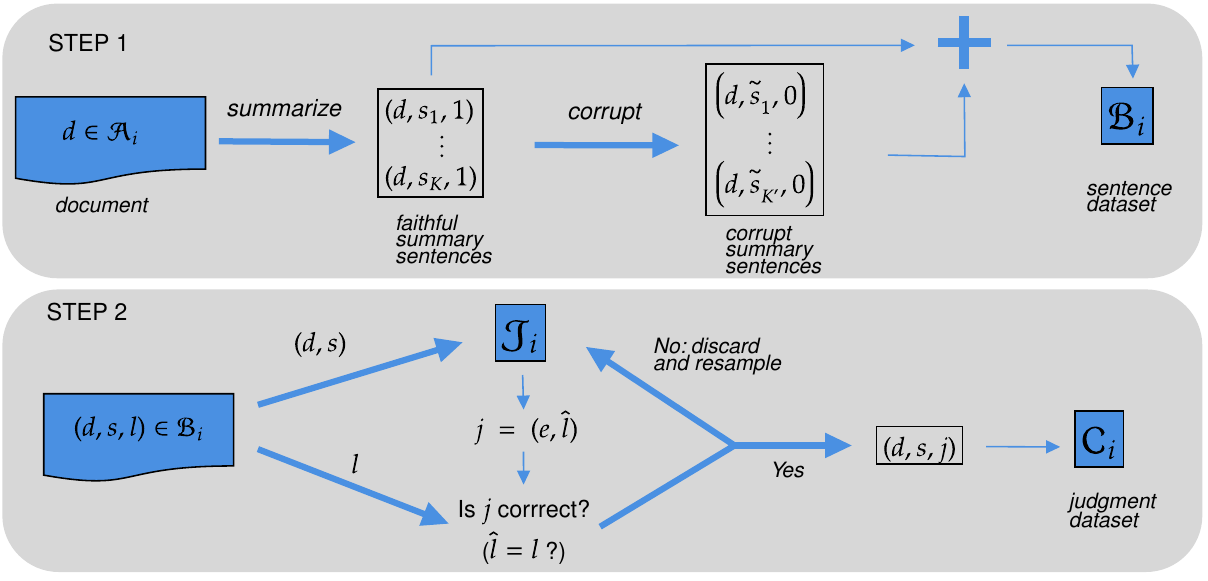}
        \caption{Our framework for self-taught multilingual evaluators \method\ includes two main steps: summary generation with/out corruption (Step 1) and judgments generation (Step 2). The judgment $j=(e,\widehat{l})$, which consists of an explanation $e$ and a faithfulness prediction $\widehat{l}$, is sampled from the current evaluator $\J_i$ given the document-sentence pair $(d,s)$. If the faithfulness prediction $\widehat{l}$ agrees with the label $l$, the judgment is accepted and added to the judgment dataset $\C_i$ with the associated document-sentence pair. \looseness = -1}
    \label{fig:method}
\end{figure*}
\section{\texttt{\method} Framework}
\label{sec:STEMF}
Our objective is to build a model that, given a document and a paragraph, returns a prediction of the faithfulness of the paragraph to the document. Our implementation is based on FineSurE \citep{song-etal-2024-finesure}, which consists of providing a prompt to an instruction-tuned LLM asking it to judge the faithfulness of a summary to the original document sentence-wise, based on a rubric of faithfulness error definitions. The LLM is also asked to specify what type of faithfulness error the sentence contains and provide an explanation for its judgment. We choose this implementation because it has shown state-of-the-art performance among prompt-based methods, and it has a simple implementation.

To further simplify the evaluator implementation, instead of feeding the model the entire paragraph, we split the paragraph into sentences and ask the model to judge the faithfulness of each sentence separately. We find that this modification helps our methodology. The prompt provided to the LLM is outlined in \Cref{prompt:judge} in \Cref{app:prompts}.

\subsection{Methodology}
\label{sec:method}
The \texttt{STEMF} approach builds a self-taught faithfulness judge in four steps, which are performed sequentially and repeated for several iterations. We provide a visual description in \Cref{fig:method}. Denote by $\J_1$ the starting LLM that will be tuned through our methodology, by $\J_{i+1}$ the LLM obtained after iteration $i$, and by $\M$ an auxiliary LLM. Let $\Omega$ be a multilingual document dataset. Our goal is to generate training data points of the form $(d,s,j)$, where $j=(e,\widehat{l})$ is a judgment on the faithfulness of the sentence $s$ to the document $d\in\Omega$ and consists of an explanation $e$ and a faithfulness prediction $\widehat{l}$. All the prompts given to $\J_i$ and $\M$ are provided in \Cref{app:prompts}. For each iteration $i\geq1$, we perform the following steps.  \looseness = -1

\paragraph{Step 0. Document selection.} We build the document dataset $\A_i$ by randomly sampling one thousand documents from $\Omega$, equally distributed among selected languages. \looseness = -1

\paragraph{Step 1. Summary generation.} For each document $d \in \A_i$, we generate a (pseudo-)faithful and a (pseudo-)corrupt summary, which are split into sentences and added to the sentence dataset $\B_i$. We consider two strategies, which are summarized in \Cref{fig:summ_strat} in \Cref{app:more_explain} and explained below.
\begin{itemize}
    \item The first strategy consists in asking $\M$ to produce a faithful summary for the document and a corrupt version of the document $\widetilde{d}$, which should contain information contradicting the original document in each of its sentences. A summary for the corrupt document is obtained by asking $\M$ to produce a faithful summary. We call this the \emph{indirect} corruption strategy.
    \item The second strategy consists in asking $\M$ to produce a faithful summary consisting of sentences $(s_1,...,s_K)$. We then obtain corrupt summary sentences by asking $\M$ to introduce a specific faithfulness error in each sentence of the faithful summary ($s_k$ is corrupted into $\widetilde{s}_k$). We call this the \emph{direct} corruption strategy. \looseness = -1
\end{itemize}
Both strategies produce the document-sentence-label triplets $\{(d, s_k, 1)\}_{k=1}^{K}$ and $\{(d, \widetilde{s}_k, 0)\}_{k=1}^{K'}$, which are added to the sentence dataset $\B_i$.

\paragraph{Step 2. Judgment generation.}
For each document-sentence-label triplet in $\B_i$, we sample a judgment $j$ from $\J_i$, as summarized in \Cref{fig:method}. A judgment consists of faithfulness \emph{prediction} $\widehat{l}$, i.e., whether the sentence $s$ is faithful to the document $d$, and an \emph{explanation} $e$ of the prediction. If the prediction is correct according to the pseudo-labels generated in Step 1, the document-sentence-judgment triplet $(d,s,j)$ is added to the judgment dataset $\C_i$. If the judgment is incorrect, a new judgment is sampled and the same process is repeated. We sample a new judgment up to two times.\looseness = -1

\paragraph{Step 3. Fine-tuning}
The last step is to fine-tune $\J_i$ on the judgment datasets $\C_i$, obtaining $\J_{i+1}$. For this step, we used supervised fine-tuning (SFT) with low rank adapters (LoRA)~\cite{hu2022lora}. 

\vspace{0.3cm}
When $\J_1$ is initialized with $\M$, the final evaluator $J_r$ (assuming $r$ iterations) is called \emph{self-taught}, as it is trained on data produced by itself. In practice, since $\M$ remains static throughout the procedure, we generate all faithful summaries, corrupt documents and corrupt summaries once and take one thousand samples at each iteration. The algorithm is summarized in Appendix~\ref{app:more_explain}. \looseness = -1

\subsection{Variations}
Besides the above \textbf{base} methodology, we propose three fine-tuning variations to inform practitioners on which strategies are effective in improving the final performance.\looseness =-1
\begin{enumerate}
    \item \textbf{Only central layers.} The first variation modifies the \emph{fine-tuning} step and consists in freezing the first and last 25\% of the layers of the LLM and only training the central layers. Following the intuition from \citet{wendler-etal-2024-llamas} and \citet{tang-etal-2024-language}, we posit that training only central layers of an LLM improves the LLM's capability at solving a particular task, without impacting its multilingual understanding and generation capabilities.
    \item \textbf{Proxy data.} Our second variation explores the effectiveness of using related data, natural language inference (NLI) data in our case, to the dataset used in the \emph{fine-tuning} step (Step 3). We include 20k examples from the multilingual dataset XNLI \citep{conneau2018xnli}, formatting them as described in \Cref{app:prompt_xnli}.
    \item \textbf{Human labels.} Lastly, we consider the setting where half of the document-sentence-label triplets in the sentence dataset $\B_i$ have been labeled faithful or unfaithful by a human.
    This addresses the issue of judgment quality assurance, as we cannot guarantee the correctness of faithful summaries or the unfaithfulness of corrupt summaries when they are generated by $\M$ in the base methodology. \looseness = -1
\end{enumerate}

\begin{table*}[!t]
\centering
\setlength{\tabcolsep}{3.9pt}
\begin{tabular}{lccccccccc}
\toprule
                           & English & French & German & Hindi & Spanish & Arabic & Italian & English& \\ \midrule
                           & \multicolumn{5}{c}{\dataname} & mFACE & \multicolumn{2}{c}{FRANK} &Avg\\ 
                           \midrule
\multicolumn{5}{l}{\emph{Models trained with our method \method}} \\
Qwen-2.5-7B-Instruct      & 69.6    & 63.5   & 57.9   & 57.5  & 61.6    & 64.2   & 68.4  &  75.0&65.1 \\
\hspace{0.2cm} + \method  & 77.8    & 71.5   & 69.5   & \underline{81.3}  & \underline{77.7}    & 72.7   & 82.6  &  86.0 &77.4\\
gemma-2-9b-it              & 76.3    & 72.1   & 67.9   & 69.1  & 73.8    & 72.8   & 81.5   &85.5 & 75.6\\
\hspace{0.2cm} + \method   & 78.4    & \underline{74.3}   & 70.6   & \textbf{81.4}  & 77.0    & 72.0   & \underline{85.0}&   \underline{86.8}& \underline{78.5}\\\midrule
\multicolumn{5}{l}{\emph{SOTA}} \\
MiniCheck-7B       & \underline{79.1}    & 66.1   & \underline{71.5}   & 73.0  & 73.2    & \underline{73.4}    & 81.6 & 84.6& 76.1\\ 
Qwen2.5-72B-Instruct       & \textbf{79.6}    & \textbf{75.6}   & \textbf{76.1}   & 76.7  & \textbf{81.0}    & \textbf{76.7}   & \textbf{85.6}   & \textbf{87.5}& \textbf{80.0}\\
\bottomrule
\end{tabular}
\caption{Balanced accuracy (in percentage) of our best trained evaluators, compared to the starting LLMs and strong baselines. The evaluations on English, French, German, Hindi, and Spanish are done on the \dataname\ dataset, while the evaluations for Arabic and Italian are done on the mFACE and FRANK datasets, respectively. We also report the performance on the original English FRANK. The Avg column reports the balanced accuracy averaged over all datasets and languages. We report in bold and underlined the highest and second highest values, respectively, for each column.}
\label{tab:best_eval}
\end{table*}

\section{Experimental Setup}
\label{sec:exp}
In this section, we describe our experimental setup. We start by discussing datasets and benchmarks, then follow by outlining our experiments.

\paragraph{Datasets for training} For all self-supervised experiments, $\Omega$ is the wikilingua dataset \cite{ladhak-etal-2020-wikilingua}, which is a collection of {\em wikihow} articles from multiple languages. The languages we consider in our investigation are German, English, Spanish, French, Hindi, Arabic, and Italian. To understand the impact of supervision and implement the third variation to the base methodology, we use human labeled data from the FRANK dataset \citep{pagnoni-etal-2021-understanding}, which contains English articles from CNN/DM \citep{hermann2015teaching} and XSum \citep{narayan-etal-2018-dont} together with LLM-generated summaries annotated sentence-wise for faithfulness. We machine translate FRANK to all the other languages we are considering, apart from Hindi since we found, through manual inspection, the translation to have low quality. \looseness = -1

\paragraph{Evaluation} We evaluate the trained LLMs on summarization and retrieval-augmented generation (RAG) benchmarks, using the balanced accuracy metric, whose definition is recalled in \Cref{app:bacc_def}. The first two benchmarks are the \dataname\ \cite{blandón2025memerag} and the mFACE \cite{aharoni-etal-2023-multilingual} datasets, which contain native multilingual faithfulness data. \dataname\ is a dataset of question-answer-label triplets, where the answer is generated by an LLM using supporting wiki articles and is labeled by a human according to its faithfulness to the supporting articles. mFACE contains articles-summary-label triplets, where the article is taken from XL-sum \citep{hasan-etal-2021-xl}, the summary is generated by an LLM, and the label says whether the summary is faithful to the article or not. We also test on the translated FRANK dataset. For details on each benchmark, see \Cref{tab:datasets} in the Appendix. \looseness = -1

\paragraph{Experiments} We perform extensive testing of our methodology. Firstly, we explore the impact of the languages included in $\A_i$ and of the starting model $\J_1$, when using the indirect corruption strategy and setting $\M$ as the starting LLM $\J_1$. We trained each starting model on each of the first five considered languages, on a mix of all seven languages (\emph{all}), and on a mix of the five European languages (\emph{euro5}). The starting models are: (i) Qwen2.5-3B-Instruct, {Qwen2.5-7B-Instruct}\footnote{https://huggingface.co/Qwen}, (ii) {gemma-2-2b-it}, {gemma-2-9b-it}\footnote{https://huggingface.co/google}, (iii) {Mistral-Nemo-Instruct-2407}\footnote{https://huggingface.co/mistralai}. We then test the direct corruption strategy and the three variations proposed in \Cref{sec:method}. For these last experiments, we use different combinations of training languages and the \texttt{Qwen2.5-7B-Instruct} model, which was the best performing model that fit in our computational setup. Lastly, a common and simple approach to tackle multilinguality is machine translating (MT) target languages to English and running inference using the English MT data, eg \cite{artetxe-etal-2023-revisiting}, which we test here. Implementation details and hyperparameters are reported in \Cref{app:hyper}.

\begin{figure*}[t]
    \centering
    \includegraphics[width=1\linewidth]{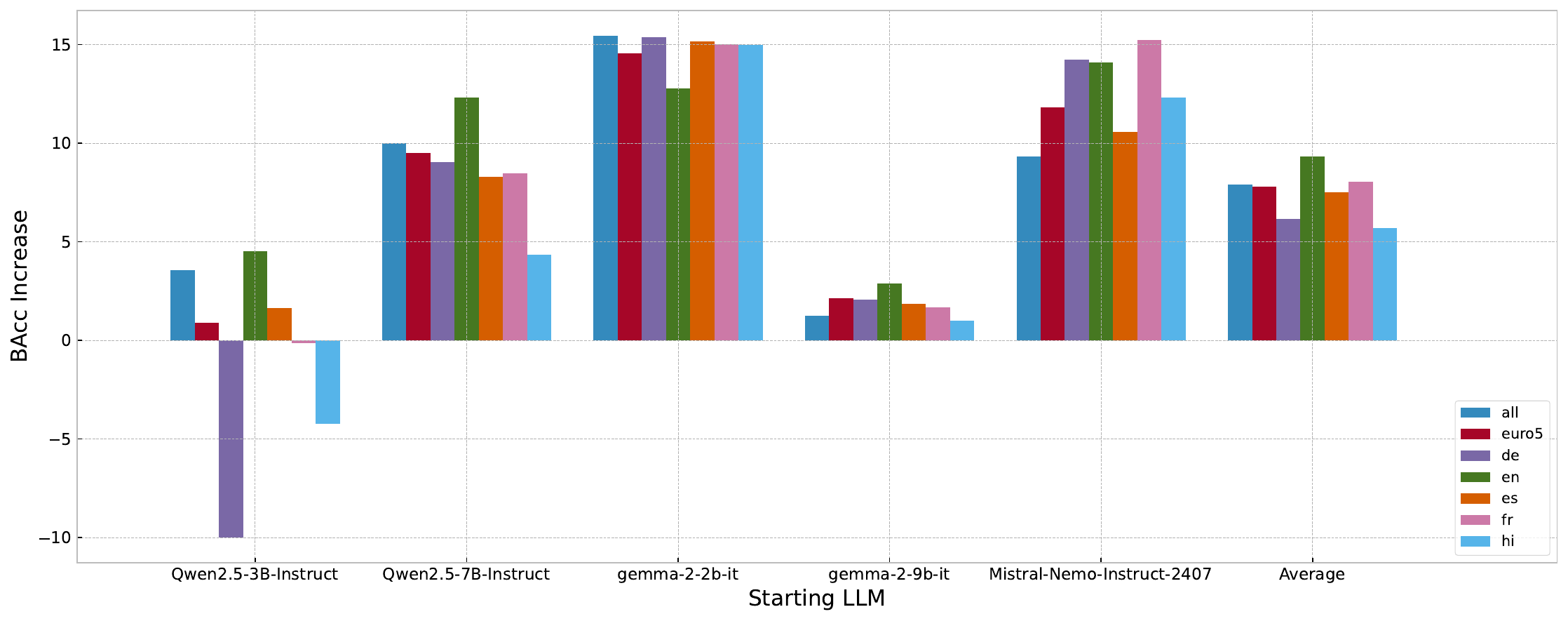}
    \caption{A comparison of training languages combinations across various LLMs. The plot reports the \emph{improvement in balanced accuracy} over the base model for each language combination used in training, in percentage points. The balanced accuracy (y-axis) is averaged over all benchmarks and languages. The models are trained using the base methodology and the indirect corruption strategy.}
    \label{fig:delta-bacc}
\end{figure*}

\section{Results}
\label{sec:res}
We provide a discussion of our results. We start by comparing the performance of our best trained evaluators against strong baselines, namely \texttt{Bespoke-MiniCheck-7B} \citep{tang2024bespokeminicheck}, a SOTA faithfulness evaluator trained on English data only, and \texttt{Qwen2.5-72B-Instruct}, a best in its class on multilingual tasks \cite{qwen2025qwen25technicalreport}\footnote{At the time of this work development, as compared to similar sized models - Llama3.1-70B, Mistral-Large, GPT4o-mini. \url{https://huggingface.co/Qwen}.}. Later in this section we discuss the results of the experiments outlined in \Cref{sec:exp}. \looseness = -1

Our best trained evaluators are \texttt{gemma-2-9b-it} and \texttt{Qwen2.5-7B-Instruct}, trained with the base STEMF framework, on exclusively English data, with the indirect corruption strategy. Their balanced accuracy is reported in \Cref{tab:best_eval} against baselines and the starting models. Both trained models outperform \texttt{Bespoke-MiniCheck-7B}, in terms of average balanced accuracy, and achieve results closer to those of \texttt{Qwen2.5-72B-Instruct}, a model eight times larger than \texttt{gemma-2-9b-it}. We also note that our trained version of \texttt{gemma-2-9b-it} has a sentence-wise balanced accuracy of 87.7\% on FRANK (English), against 86.4\% reported for GPT-4 by \citet{song-etal-2024-finesure} and 87.5\% achieved by \texttt{Qwen2.5-72B-Instruct}. To ensure these results are reproducible, we perform two additional training runs for \texttt{gemma-2-9b-it} and \texttt{Qwen2.5-7B-Instruct} and report the results \Cref{tab:best_eval_avg} in \Cref{app:complete_results}. \looseness = -1

\begin{figure*}
    \centering
    \includegraphics[width=0.99\linewidth]{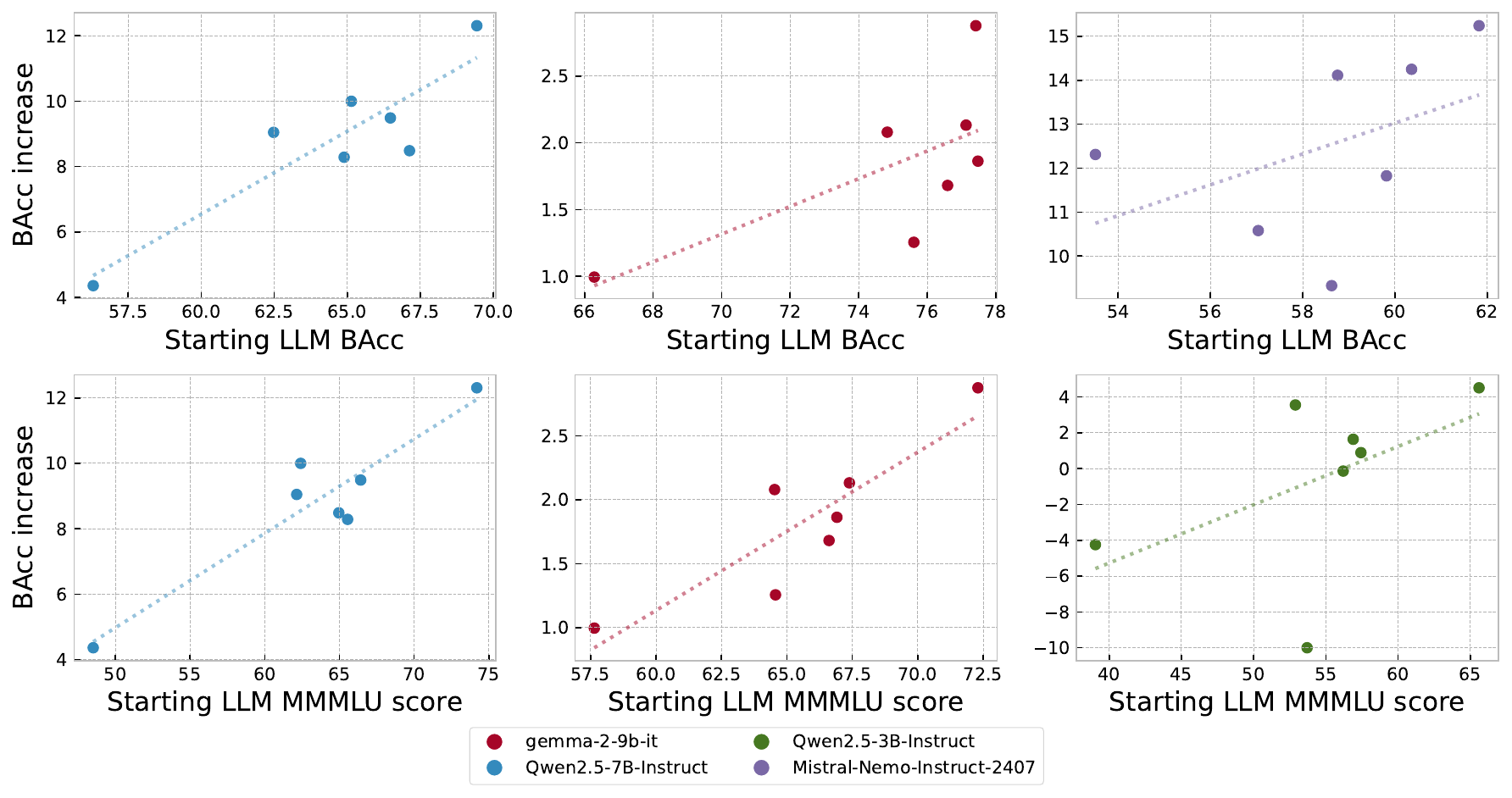}
    \caption{Relationship between performance improvement and performance of the starting LLM across the different training languages. For each trained model (represented as a dot), the plots report the improvement in averaged balanced accuracy over the starting model (y-axis) against the balanced accuracy (top x-axis) and the MMMLU score (bottom x-axis) of the starting model, averaged over the languages used in training. Each plot is specific to an LLM and present a regression line of the balanced accuracy increase against either the starting LLM balanced accuracy (top) or the starting LLM MMMLU score (bottom).}
    \label{fig:bacc_vs_base_bacc_and_mmmlu_score}
\end{figure*}
\subsection{Impact of language}
\label{sec:lang_impact}
We investigate the impact of training languages on evaluators trained with the base \method framework and the indirect corruption strategy. We report the improvement in balanced accuracy over the starting model of each trained evaluator, averaged over all test languages and benchmarks, in \Cref{fig:delta-bacc}. The exact improvement and final balanced accuracy are provided in the Appendix in Tables \ref{tab:lang_bacc_delta} and \Cref{tab:lang_bacc_final}, respectively. Our first observation is that training on exclusively English data leads to better results, on average, than any other combination of languages. As shown in \Cref{fig:delta-bacc}, evaluators trained on English present a 7.8\% average increase in balanced accuracy, while the second highest average improvement is 6.2\% and is for evaluators trained on the \emph{all} and \emph{euro5} language mixes. The lowest average improvement is 4.4\% and is for evaluators trained on Hindi. We also observe that there is no significant difference in average balanced accuracy increase (p-value $>$ 0.1) between training on a mixture of languages and training on a single language. Aside from the impact of language, these experiments also show that weaker models, such as \texttt{gemma-2-2b}, benefit largely from our training framework while the improvements are lower for the stronger counterpart, i.e.\ \texttt{gemma-2-9b-it}. \looseness = -1

These results, together with the observations on the performance of the starting models made in the previous section, suggest that \textbf{the improvements brought by our methodology are proportional to the performance of the starting model in the training languages}. We confirm this behaviour in \Cref{fig:bacc_vs_base_bacc_and_mmmlu_score}, where we plot for some of our trained models the increase in average balanced accuracy against the balanced accuracy of the starting model and the MMMLU\footnote{https://huggingface.co/datasets/openai/MMMLU, human translation of the MMLU dataset.} scores of the starting model, both averaged across the languages used in training. The MMMLU scores of all starting models are reported in \Cref{tab:mmmlu_scores} in the Appendix. We find that \textbf{including the test language among the training languages is beneficial only if the starting model is proficient enough on that test language}. We see a correlation of 0.20 (p-value $<$ 0.01) between the balanced accuracy of the trained evaluator and the presence of the test language in the training set, provided the MMMLU score of the starting model on the test language is higher than 65\%. In contrast, the correlation between the balanced accuracy of the trained evaluator and the presence of the test language in the training set, provided the MMMLU score of the starting model on the test language is lower than 65\%, is -0.21 (p-value $<$ 0.01). Similarly, the correlation between the increase in balanced accuracy of the trained evaluator and the presence of the test language in the training set, provided the MMMLU score of the starting model on the test language is lower than 65\%, is -0.18 (p-value $<$ 0.01).\looseness = -1

Next, we investigate the impact of the number of training iterations on the final model. We report in \Cref{tab:iterations} in the Appendix the average improvement in balanced accuracy for all models trained on English, at different iterations. We observe that \textbf{the third and fifth iterations are better than the first, but there is no clear difference between running the methodology for 3 or 5 iterations}. \looseness =-1

Our investigation reveals that the set of languages included in the document dataset $\A_i$ has a significant influence on the final performance of the trained evaluator. In particular, our takeaway from this first batch of experiments is that it is important to train the starting model only on languages that it is fluent in (as measured by the MMMLU performance), regardless of the test language. Moreover, we observe that training on only English data is a good choice.

\subsection{Corruption strategy}
\begin{table}[!ht]
    \centering
    \begin{tabular}{cc||cc}
        \toprule
            $\M$         & \multicolumn{1}{l||}{\texttt{Qwen-72B}}  & \multicolumn{2}{c}{\texttt{Qwen-7B}} \\ \midrule
            Corr. strat. & {direct}    & direct                           & indirect         \\ \midrule
            all          & 9.5                          & 3.9       & 10.0     \\
            euro5        & 9.4                          & 7.6       & 9.5      \\
            en           & 10.0                         & 10.2      & 12.3     \\
            Average      & 9.6                          & 7.2       & 10.6     \\ \bottomrule
        \end{tabular}
    \caption{Improvement in balanced accuracy for Indirect vs direct data synthesis strategies. We use \texttt{Qwen2.5-7B-Instruct} as the base model, and try different combinations of training languages and auxiliary model $\M$. The balanced accuracy is averaged over all languages and benchmarks.}
    \label{tab:corr_strats}
\end{table}
To test which corruption strategy is more effective, we trained the \texttt{Qwen2.5-7B-Instruct} model with the direct and indirect corruption strategies. Regarding the languages used in training, we consider the \emph{all}, \emph{euro5} and English only combinations. We initially noted that \texttt{Qwen2.5-7B-Instruct} struggles inserting faithfulness mistakes to generate corrupted summary sentences, hence we added an experiment for the direct corruption method using \texttt{Qwen2.5-72B-Instruct}. The improvement in averaged balanced accuracy over the starting model is reported in \Cref{tab:corr_strats}. When using \texttt{Qwen2.5-7B-Instruct} as $\M$, we see that the direct corruption strategy is less effective than the indirect one, especially when including languages different from English. When using \texttt{Qwen2.5-72B-Instruct} as $\M$ for the direct strategy, the difference between the performance of the two strategies gets smaller, but the indirect strategy is still uniformly better. We conclude that \textbf{the indirect corruption strategy is simpler to execute and can be performed by a relatively smaller LLM. On the contrary, the direct corruption strategy is more complex, especially in a multilingual context, and requires a large model with good multilingual capabilities}. \looseness =-1

\begin{table*}[t]
\centering
\begin{tabular}{lcccc}
\toprule
        & \multicolumn{2}{c}{Iteration 1}    & \multicolumn{2}{c}{Iteration 5}    \\ \hline
        & Full model   & Only central layers & Full model   & Only central layers \\ \hline
all     & 3.4 (0.4)  & 5.6 (6.5)        & 10.0 (10.5)  & 9.6 (10.8)        \\
en-it   & 1.9 (1.4)  & 2.9 (3.8)         & 9.9 (11.4)  & 11.6 (11.7)         \\
en      & 7.8 (7.2) & 7.5 (7.9)        & 12.3 (13.6) & 12.1 (12.7)        \\
Average & 4.4 (3.0)  & 5.4 (6.1)        & 10.7 (11.8)  & 11.1 (11.7)        \\ \hline
\end{tabular}
\caption{Comparison of full model training against training only central layers, at iterations 1 and 5. The table reports the improvement in balanced accuracy averaged over all languages and all benchmarks (averaged only over the \dataname\ dataset), for different combinations of languages used in training.}
\label{tab:only_central_layers}
\end{table*}

\subsection{Variations}
\paragraph{Training Central Layers}
We trained the central layers of the \texttt{Qwen2.5-7B-Instruct} model for three different combinations of languages used in training and compared the obtained evaluators with fully trained models on the same language combinations. We report the results in \Cref{tab:only_central_layers}, which shows that, after five iterations, there is no significant difference between the two strategies. However, the performance of the evaluators for which we only trained the central layers is higher in the first iteration, especially on the \dataname\ dataset. This behavior suggests that \textbf{training only the central layers leads to a faster convergence of \texttt{STEMF}}. Regarding computational savings, initializing and training LoRA adapters only for central layers requires half the memory of doing the same for the whole model. However, the costs of keeping the frozen full model in memory, as well as the inference costs once the adapters are merged with the full model, stay the same. Given our results and the reduced computational cost of training only the central layers of an LLM, we recommend doing so in practice. \looseness = -1

\paragraph{Proxy data} We report the performance of the STEMF framework with proxy data, compared to the performance of the base framework, in \Cref{tab:xnli} in the Appendix. Our approach to including XNLI data in the training dataset did not bring improvements, on the contrary, it hurts performance in two out of three language combinations we tried. This is in contrast to previous observations, as \citet{tang-etal-2024-minicheck} found that NLI was helpful to their methodology to train a binary classifier.

\paragraph{Human labels} We replace 50\% of the synthetic pseudo labeled data generated in Step 1 (sentence dataset $\B_i$) with human-labeled data for faithfulness from the FRANK dataset. We perform the experiment to quantify the quality of our pseudo-labels and whether human labels are needed. We find that training with human-labeled data on the \emph{all} language combination leads to an improvement in balanced accuracy of 8.0\%, against a 10\% improvement obtained by the base methodology relying on pseudo-labels only. This result indicates that synthetic data generated from the \texttt{wikilingua} dataset is sufficient for our purposes, and seems to enable better generalization of the trained model to the other datasets, i.e.\ \dataname\ and mFACE. \looseness =-1

\subsection{Baselines with Translation}
\label{sec:trans}
To understand whether translation can help in evaluating faithfulness, we re-evaluated some of the starting models on the \dataname\ dataset, but translating everything to English. The change in balanced accuracy is reported in \Cref{tab:translation} which shows that translation often hurts performance, with the exception of Hindi where translation brings improvements almost everywhere, probably due to the low proficiency of the models in this language.
\begin{table}[ht]
    \centering
    \begin{tabular}{lcccc}
    \toprule
                               & de & es & fr &  hi \\ \midrule
    Mistral-Nemo-Instruct &   -1.6 &     2.6 &    0.5 &    0.7 \\
    Qwen2.5-3B-Instruct        &   -2.5 &    -1.5 &   -5.2 &    4.1 \\
    Qwen2.5-7B-Instruct        &    0.3 &    -3.0 &   -3.3 &    2.6 \\
    gemma-2-9b-it              &   -2.6 &    -5.4 &   -1.8 &    1.4 \\ \bottomrule
    \end{tabular}
    \caption{Evaluators results comparing using the target language vs. a ``pivot'' approach of translating into English first. We report the difference in balanced accuracy, for different starting models, on the \dataname\ dataset. \looseness = -1}
    \label{tab:translation}
\end{table}

\section{Conclusion}
We presented \method, a framework to train LLM-based multilingual faithfulness evaluators through synthetic data, which led to substantial improvements across most starting LLMs, with an average increase of $6.9\%$  in balanced accuracy. Our best evaluator, based on \texttt{gemma-2-9b-it} and trained on English data, achieves competitive performance with much larger models and even outperforms GPT-4 on the FRANK benchmark. We identified several key insights for developing effective evaluators across languages. Our results indicate that practitioners should prioritize selecting starting models with strong proficiency in their target languages when performing self-taught approaches, with training on English often yielding the best results across all test languages. Additionally, we found that training only the central layers of an LLM achieves comparable results to training all layers while being more computationally efficient. Lastly, we found that translation-based approaches generally degrade performance except for languages where the LLM has a very limited proficiency. \looseness = -1

\section*{Limitations}
Our \texttt{STEMF} methodology is primarily focused on faithfulness evaluation. In practice however, there exists several other dimensions that are required to get a comprehensive evaluation of LLM outputs in the settings of summarization and QA, e.g. informativeness, harmfulness and fluency. We leave this investigation for future work. 
Under the experimental conditions herein described, our results show that training on English is sufficient. We hypothesize that this relates to strong general language capability on English in comparison to other languages, and a language agnostic representation of faithfulness features, which is left for future work.
Finally, our study would not have been possible without the faithfulness benchmarks covering the selected languages. We believe that it is crucial to extend these benchmarks, as well as the automatic evaluation research, to low-resource languages in order to ensure the usefulness and safety of LLM-generated content across all cultures.

\bibliography{anthology,custom}
\clearpage
\appendix

\section{Evaluators built with synthetic data}
\label{app:comparison}
In this section, we compare our \method\ framework to those designed by \citet{wang2024self} and \citet{tang-etal-2024-minicheck}, who focused exclusively on English.

\citet{wang2024self} build a self-taught evaluator which, given a generic instruction and two responses, ranks the two responses after providing an explanation. They follow a procedure similar to that outlined in \Cref{fig:method}, but provide both the good (faithful) and the bad (corrupt) response to the judge and only accept the judgment if it ranks the good response first. In the context of this work, the faithfulness of a summary is a binary variable, which is why we design STEMF to evaluate single summaries and avoid comparisons.

\citet{tang-etal-2024-minicheck} build a binary classifier which, given a document and a claim, judges whether the claim is faithful to the document but does not provide an explanation. Similarly to us, they build synthetic faithful and corrupt data, but use a different approach. They produce synthetic data in two ways. The first (Claim to Doc) starts with a claim, decomposes it into atomic facts, and then uses LLMs to generate both supporting and non-supporting documents. The second (Doc to Claim) begins with human-written documents, summarizes them into claims, and then augments this data by modifying/clipping documents or pairing claims with different document sections. Differently from us, their judgment dataset does not include explanations, which could instead help in interpreting predictions.

\section{Prompts}
\label{app:prompts}
\subsection{Evaluator prompt}
The prompt used by our evaluators is provided in \Cref{prompt:judge}.

\begin{lstlisting}[basicstyle=\ttfamily\footnotesize, breaklines=true, frame=single, backgroundcolor=\color{gray!10}, caption=Prompt for the faithfulness evaluator, label=prompt:judge]
You will receive a text followed by a statement. Your task is to assess the factuality of the statement with respect to the source text across nine categories:
* no error: the statement aligns explicitly with the content of the text and is faithful to it.
* out-of-context error: the statement contains information not present in the text.
* entity error: the primary arguments (or their attributes) of the predicate are wrong.
* predicate error: the predicate in the statement is inconsistent with the text.
* circumstantial error: the additional information (like location or time) specifying the circumstance around a predicate is wrong.
* grammatical error: the grammar of the statement is so wrong that it becomes meaningless.
* coreference error: a pronoun or reference with wrong or non-existing antecedent.
* linking error: error in how multiple statements are linked together in the discourse (for example temporal ordering or causal link).
* other error: the statement contains any factuality error which is not defined here.

Instruction:
First, compare the statement with the text.
Second, provide a single sentence explaining which factuality error the statement has.
Third, answer the classified error category for the statement.

Provide your answer in JSON format. The answer should be a dictionary whose keys are "reason", and "category":
{"reason": "your reason", "category": "no error"} or {"reason": "your reason", "category": "which error"}

Text:
<replace text here>
Statement:
<replace statement here>
\end{lstlisting}

\subsection{Summary generation prompts}
The prompt used to generate a faithful summary of a given article is given in \Cref{prompt:good_summ}.
The prompts used to corrupt each sentence in a summary according to the direct corruption strategy depend on the selected type of faithfulness error and are provided in \Cref{prompt:direct1} (predicate error), \Cref{prompt:direct2} (entity error), \Cref{prompt:direct3} (circumstantial error), \Cref{prompt:direct4} (linking error), \Cref{prompt:direct5} (out-of-context error). Lastly, the prompt used to ask for a corrupted version of an article is given in \Cref{prompt:corrupt_article}.

\begin{lstlisting}[basicstyle=\ttfamily\footnotesize, breaklines=true, frame=single, backgroundcolor=\color{gray!10}, caption=Prompt to ask for a faithful summary, label=prompt:good_summ]
You will be provided with an article containing instructions to complete a task or deal with a situation. Please provide a concise and faithful summary for the article. Provide the summary as a list of sentences separated by the characters '###'. That is,

### First sentence.
### Second sentence.
### Third sentence.
and so on.

Article:
<replace article here>
\end{lstlisting}

\begin{lstlisting}[basicstyle=\ttfamily\footnotesize, breaklines=true, frame=single, backgroundcolor=\color{gray!10}, caption=Prompt to insert a predicate error in a sentence, label=prompt:direct1]
You will be provided with a sentence and a source text. First, individuate the main clause in the sentence. Then, individuate the subject, the predicate, the object and the attributes of the main clause. Your task is to modify the predicate and/or the object of the main clause so that it is inconsistent with the orignal one and the source text. Keep the subject and the attributes similar to the original sentence. Provide your answer in the following format.

### Original sentence: <original sentence>
### Main clause: <main clause>
### Subject: <subject>
### Predicate: <predicate>
### Object: <object>
### Attributes: <attributes>
### Strategy: <how you are going to modify the sentence>
### <modified sentence>

Do not provide additional text after the modified sentence.

Sentence:
<replace sentence here>

Source text:
<replace text here>
\end{lstlisting}

\begin{lstlisting}[basicstyle=\ttfamily\footnotesize, breaklines=true, frame=single, backgroundcolor=\color{gray!10}, caption=Prompt to insert an entity error in a sentence, label=prompt:direct2]
You will be provided with a sentence and a source text. First, individuate the main clause in the sentence. Then, individuate the subject, the predicate, the object and the attributes of the main clause. Your task is to modify the subject of the main clause so that it is inconsistent with the original one and the text. Keep the predicate, the object and the attributes similar to the original sentence. Provide your answer in the following format.

### Original sentence: <original sentence>
### Main clause: <main clause>
### Subject: <subject>
### Predicate: <predicate>
### Object: <object>
### Attributes: <attributes>
### Strategy: <how you are going to modify the sentence>
### <modified sentence>

Do not provide additional text after the modified sentence.

Sentence:
<replace sentence here>

Source text:
<replace text here>
\end{lstlisting}

\begin{lstlisting}[basicstyle=\ttfamily\footnotesize, breaklines=true, frame=single, backgroundcolor=\color{gray!10}, caption=Prompt to insert a circumstantial error in a sentence, label=prompt:direct3]
You will be provided with a sentence and a source text. First, individuate the main clause in the sentence. Then, individuate the subject, the predicate, the object and the attributes of the main clause. Your task is to modify the attributes (e.g. location, time, manner, direction, modality) of the main clause so that it is inconsistent with the original one and the text. Keep the subject, the predicate, and the object similar to the original sentence. Provide your answer in the following format.

### Original sentence: <original sentence>
### Main clause: <main clause>
### Subject: <subject>
### Predicate: <predicate>
### Object: <object>
### Attributes: <attributes>
### Strategy: <how you are going to modify the sentence>
### <modified sentence>

Do not provide additional text after the modified sentence.

Sentence:
<replace sentence here>

Source text:
<replace text here>
\end{lstlisting}

\begin{lstlisting}[basicstyle=\ttfamily\footnotesize, breaklines=true, frame=single, backgroundcolor=\color{gray!10}, caption=Prompt to insert a linking error in a sentence, label=prompt:direct4]
You will be provided with a sentence and a source text. First, analyze the sencence and individuate its clauses. Then, modify the sentence so that the temporal ordering or the discourse links (e.g. RST relations, discourse connectors) among its clauses are inconsistent with the original sentence and the text. Provide your answer in the following format.

### Original sentence: <original sentence>
### First clause: <first clause>
### Second clause: <second clause>
### ...
### Strategy: <how you are going to modify the sentence>
### <modified sentence>

Do not provide additional text after the modified sentence.

Sentence:
<replace sentence here>

Source text:
<replace text here>
\end{lstlisting}
\begin{figure*}[t]
    \centering
    \includegraphics[width=1\linewidth]{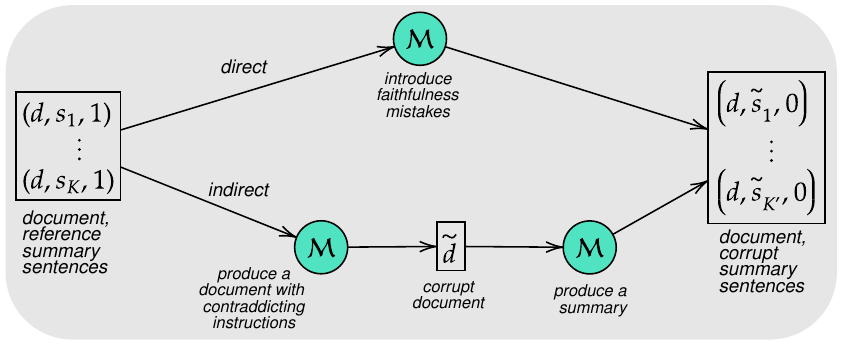}
    \caption{Scheme for producing corrupted sentences, given a document and faithful sentences. Replaces the "corrupt" arrow in \Cref{fig:method}.}
    \label{fig:summ_strat}
\end{figure*}
\begin{lstlisting}[basicstyle=\ttfamily\footnotesize, breaklines=true, frame=single, backgroundcolor=\color{gray!10}, caption=Prompt to insert an out-of-context error in a sentence, label=prompt:direct5]
You will be provided with a sentence and a source text. Your task is to modify the sentence so that it contains information on a matter not discussed in the source text. Provide your answer in the following format.

### Original sentence: <original sentence>
### Strategy: <how you are going to modify the sentence>
### <modified sentence>

Do not provide additional text after the modified sentence.

Sentence:
<replace sentence here>

Source text:
<replace text here>
\end{lstlisting}

\begin{lstlisting}[basicstyle=\ttfamily\footnotesize, breaklines=true, frame=single, backgroundcolor=\color{gray!10}, caption=Prompt to corrupt an article, label=prompt:corrupt_article]
You will be provided with an article containing instructions to complete a task or deal with a situation. The article is titled "{title}". Please provide contraddicting instructions for the same tasks. Make sure the new instrcutions are self-coherent and plausible. Maintain the same language, structure and style of the article.

Article:
<replace article here>
\end{lstlisting}

\subsection{XNLI data prompt}
\label{app:prompt_xnli}
To include XNLI datapoints within the training dataset, we use the text in \Cref{prompt:ask_xnli} as the prompt and the text in \Cref{prompt:tell_xnli} as the accepted judgment.
\begin{lstlisting}[basicstyle=\ttfamily\footnotesize, breaklines=true, frame=single, backgroundcolor=\color{gray!10}, caption=Prompt for XNLI data, label=prompt:ask_xnli]
You will be given two sentences, a premise and a hypothesis. Your task is to determine whether the premise implies, contradicts, or neither implies nor contradicts the hypothesis.

Premise: <replace premise here>
Hypothesis: <replace hypthesis here>
\end{lstlisting}

\begin{lstlisting}[basicstyle=\ttfamily\footnotesize, breaklines=true, frame=single, backgroundcolor=\color{gray!10}, caption=Accepted judgment for XNLI data, label=prompt:tell_xnli]
<Based on XNLI datapoint label, choose one of:
"The premise implies the hypothesis"
"The premise contradicts the hypothesis"
"The premise neither implies nor contradicts the hypothesis">
\end{lstlisting}

\section{Additional framework description}
\label{app:more_explain}
We visually describe the direct and indirect strategies for generating summaries in \Cref{fig:summ_strat}. We summarize the STEMF framework with the algorithm below.

\begin{algorithm}
\caption{STEMF: Self-Taught Evaluator for Multilingual Faithfulness}
\label{alg:repeat}
\begin{algorithmic}[1]
\State \textbf{Input:} Multilingual documents $\Omega$, auxiliary LLM $\M$
\State \textbf{Output:} Multilingual self-taught evaluator $\J_r$
\State $\J_0 \gets M$
\For{$i = 1$ to $r$}
    \State $\A_i \gets \text{sample}(\Omega)$ \Comment{Step 0}
    \State $\B_i \gets \emptyset$, $\C_i \gets \emptyset$
    \For{each $d \in \A_i$} \Comment{Step 1}
        \State $\{s_k\}_{k=1}^K \gets M_{\text{Summarize}}(d)$
        \For{each sentence $s_k$}
            \State $\B_i \gets \B_i \cup \{(d, s_k, 1)\}$
            \State $\widetilde{s}_k = M_\text{Corrupt}(s_k)$
            \State $\B_i \gets \B_i \cup \{(d, \widetilde{s}_k, 0)\}$
        \EndFor
    \EndFor
    \For{each $(d, s, l)$ in $\B_i$} \Comment{Step 2}
        \State $(e, \widehat{l}) \gets \J_{\text{judge}}(d, s)$
        \If{$\widehat{l} = l$}
            \State $\C_i \gets \C_i \cup (d, s, l)$
        \EndIf
    \EndFor
    \State $\J_i \gets \text{Finetune}(\J_{i-1}, \C_i)$ \Comment{Step 3}
\EndFor
\end{algorithmic}
\end{algorithm}

\section{Evaluation metrics}
\label{app:bacc_def}
We give here a reminder on the definition of balanced accuracy. Let $\A$ be a set of predictions, made of the set of true positives $TP$, the set of false positives $FP$, the set of true negatives $TN$, and the set of false negatives $FN$. Let the true positive rate be defined as $|TP| / (|TP| + |FN|)$ and the true negative rate be defined as $|TN| / (|TN| + |FP|)$. The balanced accuracy is defined as the average between the true positive rate and true negative rate. \looseness = -1

\section{Implementation details}
\label{app:hyper}
We use \texttt{vllm} \cite{kwon2023efficient} for inference and \texttt{alignment-handbook} \cite{Tunstall_The_Alignment_Handbook} for training the LLMs, adding some modifications to allow training only the central layers. Our experiments are run on 4 Nvidia H100 GPUs, when fine-tuning \texttt{gemma-2-9b}, or on 4 Nvidia L40S GPUs, in all other cases. We provide the hyperparamers used for training in \Cref{tab:hyper_train} and the hyperparameters used for inference on \texttt{vllm} in \Cref{tab:hyper_vllm}.

\begin{table}[ht]
\centering
\begin{tabular}{lc}
\toprule
    \multicolumn{1}{c}{Parameter} & Value \\ \midrule
    Number of epochs             & 1    \\
    Gradient Accumulation Step   & 16   \\
    Batch Size                   & 4     \\
    Learning rate                & 5e-5 \\
    LoRA Rank                    & 128  \\
    LoRA Alpha                   & 256  \\
    Lora Dropout                 & 0.05 \\
    Max length                   & 2048    \\
    \bottomrule
\end{tabular}
\caption{Hyper-parameter settings for LLM training.}
\label{tab:hyper_train}
\end{table}

\begin{table}[ht]
\centering
\begin{tabular}{lc}
\toprule
    \multicolumn{1}{c}{Parameter} & Value \\ \midrule
    Temperature                   & 1      \\
    Temperature (for judgments)   & 0      \\
    Max tokens                    & 4096   \\
    Top\_p                        & 0.8     \\
    \bottomrule
\end{tabular}
\caption{Hyper-parameter settings for LLM inference.}
\label{tab:hyper_vllm}
\end{table}

\section{Tables}
\label{app:complete_results}
We provide here tables with detailed results on our experiments.

\begin{table}[!ht]
    \centering
    \begin{tabular}{lcc}
    \toprule
            &   with XNLI &  without XNLI \\
    \midrule
    all     &  6.7        &  10.0 \\
    euro5   &  7.1        &  9.5  \\
    en      &  12.4       &  12.3 \\
    Average &  8.8        &  10.6 \\
    \bottomrule
    \end{tabular}
    \caption{Improvement in average balanced accuracy, in percentage, over the starting model for \texttt{Qwen2.5-7B-Instruct} trained with or without XNLI data, on three combinations of training languages. \looseness = -1}
    \label{tab:xnli}
\end{table}

\begin{table}[!ht]
    \centering
    \setlength{\tabcolsep}{5pt}
    \begin{tabular}{lccc}
        \toprule
         & 1 & 3 & 5 \\
        \midrule
        Mistral-Nemo-Instruct-2407 & 7.5 & 16.1 & 14.1 \\
        Qwen2.5-3B-Instruct & 3.7 & 4.5 & 4.5 \\
        Qwen2.5-7B-Instruct & 7.8 & 12.1 & 12.3 \\
        gemma-2-2b-it & 9.8 & 13.3 & 12.8 \\
        gemma-2-9b-it & 2.4 & 2.3 & 2.9 \\
        Average & 5.2 & 8.2 & 7.8 \\
        \bottomrule
    \end{tabular}
    \caption{Impact of the number of training iterations on model performance. The LLMs are trained using English (best language for training) and the performance is measured by absolute improvement in average balanced accuracy across languages and datasets.}
    \label{tab:iterations}
\end{table}

\begin{table*}[!t]
\centering
\begin{tabular}{lcccccc}
\toprule
                           & English & French & German & Hindi & Spanish & Arabic \\ \midrule
                           & \multicolumn{5}{c}{\dataname} & mFACE \\ 
                           \midrule
\multicolumn{5}{l}{\emph{Models trained with our method \method}} \\
Qwen-2.5-7B-Instruct      & 69.6    & 63.5   & 57.9   & 57.5  & 61.6    & 64.2   \\
\hspace{0.2cm} + \method  & 76.4$\pm$0.9    & 71.7$\pm$0.6   & 68.1$\pm$1.0   & 75.9$\pm$2.4  & 75.2$\pm$1.5    & \underline{73.5$\pm$1.1}   \\
gemma-2-9b-it              & 76.3    & 72.1   & 67.9   & 69.1  & 73.8    & 72.8   \\
\hspace{0.2cm} + \method   & 74.4$\pm$1.8    & \textbf{75.9$\pm$0.7}   & 66.3$\pm$1.9   & \textbf{81.9$\pm$0.2}  & \underline{77.6$\pm$1.2}    & 70.9$\pm$0.8   \\\midrule
\multicolumn{5}{l}{\emph{SOTA}} \\
MiniCheck-7B       & \underline{79.1}    & 66.1   & \underline{71.5}   & 73.0  & 73.2    & 73.4    \\ 
Qwen2.5-72B-Instruct       & \textbf{79.6}    & \underline{75.6}   & \textbf{76.1}   & \underline{76.7}  & \textbf{81.0}    & \textbf{76.7}   \\
\bottomrule
\end{tabular}
\begin{tabular}{lccc}
\toprule
                           & Italian & English& Avg\\ \midrule
                           & \multicolumn{2}{c}{FRANK} &\\ 
                           \midrule
\multicolumn{3}{l}{\emph{Models trained with our method \method}} \\
Qwen-2.5-7B-Instruct      & 68.4  &  75.0&65.1 \\
\hspace{0.2cm} + \method  & 81.0$\pm$1.2  &  84.0$\pm$1.3 &75.7$\pm$1.3\\
gemma-2-9b-it             & 81.5   &85.5 & 75.6\\
\hspace{0.2cm} + \method  & \underline{84.4$\pm$0.7}&   \underline{85.7$\pm$0.7}& \underline{77.5$\pm$0.8}\\\midrule
\multicolumn{3}{l}{\emph{SOTA}} \\
MiniCheck-7B              & 81.6 & 84.6& 76.1\\ 
Qwen2.5-72B-Instruct      & \textbf{85.6}   & \textbf{87.5}& \textbf{80.0}\\
\bottomrule
\end{tabular}
\caption{Same as \Cref{tab:best_eval}, but we report the average and standard error for three separate runs of STEMF. Balanced accuracy (in percentage) of our best trained evaluators, compared to the starting LLMs and strong baselines. The evaluations on English, French, German, Hindi, and Spanish are done on the \dataname\ dataset, while the evaluations for Arabic and Italian are done on the mFACE and FRANK datasets, respectively. We also report the performance on the original English FRANK. The Avg column reports the balanced accuracy averaged over all datasets and languages. We report in bold and underlined the highest and second highest values, respectively, for each column.}
\label{tab:best_eval_avg}
\end{table*}

\begin{table*}[t]
\centering
\begin{tabular}{lccccc}
\toprule
Name    & Native languages       & Translated to      & \# annotators & \# factual & \# non-factual \\ \midrule
FRANK   & en                 & ar, de, es, fr, it & 3            & 3063      & 1829          \\
MEMERAG & de, en, es, fr, hi & /                  & 5            & 986       & 314           \\
mFACE   & ar, en, es, fr, hi     & /                  & 3            & 1362      & 1931         \\ \bottomrule
\end{tabular}
\caption{Details on evaluation benchmarks. 
}
\label{tab:datasets}
\end{table*}
\begin{table*}[t]
\centering
\begin{tabular}{lccccccc}
\toprule
                    & Arabic    & German    & English   & Spanish    & French    & Hindi    & Italian    \\
\midrule
Qwen2.5-3B-Instruct  & 44.0 & 53.7 & 65.6 & 56.9 & 56.2 & 39.1 & 54.7 \\
Qwen2.5-7B-Instruct  & 56.2 & 62.1 & 74.2 & 65.6 & 65.0 & 48.5 & 65.3 \\
Qwen2.5-72B-Instruct & 74.3 & 72.5 & 86.1 & 77.5 & 76.0 & 69.1 & 72.5 \\
gemma-2-2b-it        & 36.0 & 47.0 & 56.1 & 48.0 & 47.4 & 38.3 & 46.1 \\
gemma-2-9b-it        & 57.7 & 64.5 & 72.3 & 66.9 & 66.6 & 57.6 & 66.6 \\
Mistral-Nemo-Instruct-2407 & 47.8 & 53.9 & 68.0 & 58.9 & 59.5 & 47.1 & 58.9 \\
\bottomrule
\end{tabular}
\caption{MMMLU scores for the models considered in this work.}
\label{tab:mmmlu_scores}
\end{table*}
\begin{table*}[ht]
    \centering
    \begin{tabular}{lrrrrrrr}
\toprule
{} &  all &  euro5 &    German &    English &    Spanish &    French &    Hindi \\
\midrule
Mistral-Nemo-Instruct-2407 &                   9.3 &            11.8 &  14.3 &  14.1 &  10.6 &  15.2 &  12.3 \\
Qwen2.5-3B-Instruct        &                   3.6 &             0.9 & -10.0 &   4.5 &   1.6 &  -0.1 &  -4.2 \\
Qwen2.5-7B-Instruct        &                  10.0 &             9.5 &   9.0 &  12.3 &   8.3 &   8.5 &   4.4 \\
gemma-2-2b-it              &                  15.5 &            14.6 &  15.4 &  12.8 &  15.2 &  15.0 &  15.0 \\
gemma-2-9b-it              &                   1.3 &             2.1 &   2.1 &   2.9 &   1.9 &   1.7 &   1.0 \\
Average                    &                   6.2 &             6.2 &   4.7 &   7.8 &   6.1 &   6.1 &   4.4 \\
\bottomrule
\end{tabular}
    \caption{Improvement in balanced accuracy, in percentage points, over the starting model, for the considered starting model and training languages combinations. The balanced accuracy is averaged over all benchmarks and languages. The models are trained using the base methodology and the indirect corruption strategy.}
    \label{tab:lang_bacc_delta}
\end{table*}

\begin{table*}[!ht]
    \centering
    \begin{tabular}{lrrrrrrr}
\toprule
{} &  all &  euro5 &    German &    English &    Spanish &    French &    Hindi \\
\midrule
Mistral-Nemo-Instruct-2407 &                  68.0 &            70.5 &  72.9 &  72.7 &  69.2 &  73.9 &  70.9 \\
Qwen2.5-3B-Instruct        &                  72.2 &            69.5 &  58.6 &  73.1 &  70.3 &  68.5 &  64.4 \\
Qwen2.5-7B-Instruct        &                  75.1 &            74.6 &  74.2 &  77.4 &  73.4 &  73.6 &  69.5 \\
gemma-2-2b-it              &                  72.0 &            71.1 &  71.9 &  69.3 &  71.7 &  71.5 &  71.5 \\
gemma-2-9b-it              &                  76.9 &            77.7 &  77.7 &  78.5 &  77.5 &  77.3 &  76.6 \\
Average                    &                  72.6 &            72.7 &  71.1 &  74.2 &  72.5 &  72.6 &  70.9 \\
\bottomrule
\end{tabular}
    \caption{Final balanced accuracy, in percentage, for the considered starting model and training languages combinations. The balanced accuracy is averaged over all benchmarks and languages. The models are trained using the base methodology and the indirect corruption strategy.}
    \label{tab:lang_bacc_final}
\end{table*}

\end{document}